\documentclass[10pt,twocolumn,letterpaper]{article}
\usepackage[pagenumbers]{wacv} 

\usepackage{graphicx}
\usepackage{amsmath}
\usepackage{amssymb}
\usepackage{booktabs}
\usepackage{morefloats}
\extrafloats{100}
\usepackage{etex}
\usepackage[pagebackref,breaklinks,colorlinks]{hyperref}

\usepackage[capitalize]{cleveref}
\crefname{section}{Sec.}{Secs.}
\Crefname{section}{Section}{Sections}
\Crefname{table}{Table}{Tables}
\crefname{table}{Tab.}{Tabs.}

\def\wacvPaperID{2244} 
\def\confName{WACV}
\def\confYear{2025}

\begin{document}

\title{STEAM-EEG: Spatiotemporal EEG Analysis with Markov Transfer Fields and Attentive CNNs}

\author{Jiahao Qin\\
Xi'an Jiaotong-Liverpool University\\
Suzhou, China\\
{\tt\small Jiahao.qin19@gmail.com}
\and
Feng Liu\\
Shanghai Jiao Tong University\\
Shanghai, China\\
{\tt\small lsttoy@163.com}
}
\maketitle

\begin{abstract}
   Electroencephalogram (EEG) signals play a pivotal role in biomedical research and clinical applications, including epilepsy diagnosis, sleep disorder analysis, and brain-computer interfaces. However, the effective analysis and interpretation of these complex signals often present significant challenges. This paper presents a novel approach that integrates computer graphics techniques with biological signal pattern recognition, specifically using Markov Transfer Fields (MTFs) for EEG time series imaging. The proposed framework (STEAM-EEG) employs the capabilities of MTFs to capture the spatiotemporal dynamics of EEG signals, transforming them into visually informative images. These images are then rendered, visualised, and modelled using state-of-the-art computer graphics techniques, thereby facilitating enhanced data exploration, pattern recognition, and decision-making. The code could be accessed from GitHub.
\end{abstract}

\section{Introduction}
\label{sec:introduction}

Electroencephalogram (EEG) signals play a crucial role in neuroscience, clinical diagnosis, and brain-computer interfaces due to their non-invasive nature and high temporal resolution \cite{Gu_2021_EEG,aggarwal2022review,ABDULKADER2015213,HRAMOV20211}. However, EEG signal analysis presents significant challenges owing to its complex, non-stationary nature, and the presence of noise and artifacts. Traditional approaches, relying on manual feature engineering and conventional machine learning algorithms \cite{lotte_review_2007,taheri_gorji2023using,min2023fusion,DADEBAYEV20224385}, often fail to capture the intricate spatiotemporal dynamics inherent in EEG signals. Recent advancements in deep learning, particularly convolutional neural networks (CNNs), have revolutionized this field by automatically learning hierarchical representations from raw data \cite{rajwal2023convolutional,10102335,10384632}, addressing many limitations of traditional methods.

The evolution of EEG signal analysis has seen the integration of various advanced techniques. Attention mechanisms have been introduced to focus on relevant parts of the input signal, enhancing the model's ability to capture critical information \cite{9669037,9417097,9204431}. Markov Transfer Fields (MTFs) have demonstrated promise in modeling complex spatiotemporal patterns, providing a more comprehensive representation of EEG dynamics \cite{746636,ZHU2021540,app14177720}. Additionally, Singular Spectrum Analysis (SSA) has been employed to decompose EEG signals into trend, seasonal, and noise components, facilitating more nuanced feature extraction \cite{sanei_singular_2013,7462195,s18030697,s23031235,MA2022108679}.

\begin{figure}[!t]
\centering
\includegraphics[width=0.9\columnwidth]{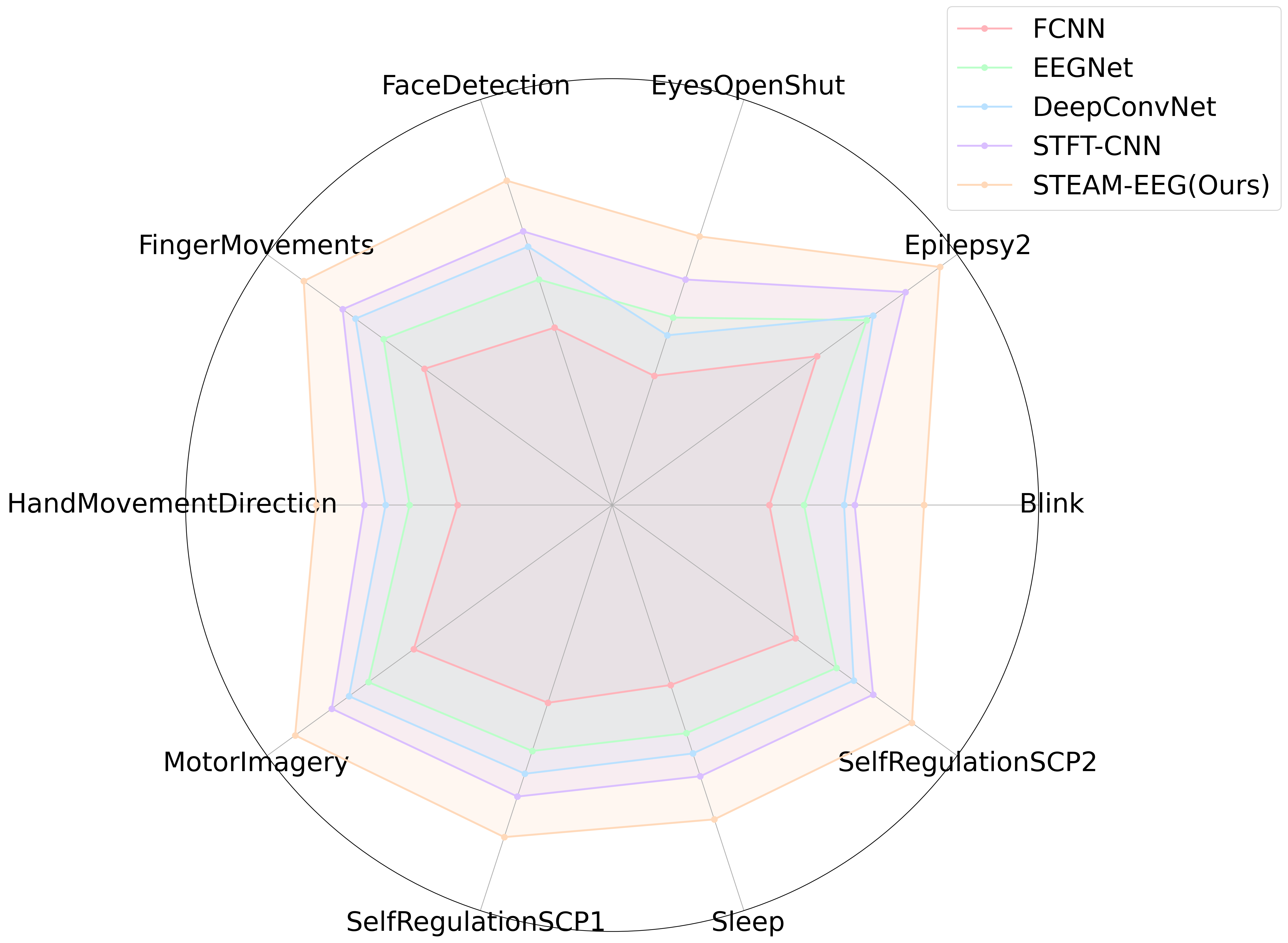}
\caption{Performance comparison of our proposed SOTA method against baseline approaches across various EEG datasets. The radar chart illustrates the superior accuracy of our method (in orange) compared to FCNN, EEGNet, DeepConvNet, and STFT-CNN across ten different EEG datasets.}
\label{fig:performance_comparison}
\end{figure}

Despite these advancements, there remains a need for a comprehensive approach that effectively integrates these diverse techniques to address the multifaceted challenges of EEG signal analysis. To address this gap, we propose a novel framework that combines Trend-Seasonal Decomposition using SSA, parallel 1D-CNNs with modified attention mechanisms, and MTF imaging. This approach integrates advanced signal processing, deep learning, and computer graphics techniques for enhanced EEG signal analysis. Figure \ref{fig:performance_comparison} illustrates the performance of our proposed method compared to existing state-of-the-art approaches across a diverse range of EEG datasets. The radar chart clearly demonstrates the superior accuracy of our method across all evaluated datasets, highlighting the effectiveness of our integrated approach in capturing and analyzing the complex spatiotemporal dynamics inherent in EEG signals. Our main contributions are: 
\begin{itemize}
    \item We propose a a novel integrated framework (STEAM-EEG) for EEG signal analysis that synergistically combines Singular Spectrum Analysis (SSA) for trend-seasonal decomposition, parallel 1D-CNNs with a modified attention mechanism, and Markov Transfer Field (MTF) imaging. This unique combination allows for more effective capture and analysis of complex spatiotemporal dynamics in EEG signals. 
    \item We introduce a modified attention mechanism specifically designed to capture cross-channel dependencies in parallel 1D-CNNs, enhancing the model's ability to focus on relevant features across multiple EEG channels. 
    \item We leverage MTF imaging to model spatiotemporal dynamics and generate informative visual representations of EEG patterns, improving both the analysis and interpretability of the results. 
    \item We conduct extensive evaluations on diverse EEG datasets, demonstrating significant improvements in accuracy and robustness compared to state-of-the-art methods across various EEG analysis tasks.
\end{itemize}

\section{Related Work}
\label{sec:related_work}

EEG signal analysis has evolved from traditional time-domain and frequency-domain techniques \cite{KHOSLA2020649,WAN20211, Hatamoto2022SleepEA} to more sophisticated machine learning approaches. Support vector machines, k-nearest neighbors, and decision trees have shown promise in various EEG classification tasks \cite{rajwal2023convolutional,10102335,10384632}, enhancing diagnostic accuracy \cite{subasi_epileptic_2010, zhang_multimodal_2017}. Recent years have seen a shift towards deep learning, particularly convolutional neural networks (CNNs). These models have achieved state-of-the-art performance in seizure detection \cite{rajwal2023convolutional,10102335}, emotion recognition \cite{Liu_2020_eeg,DADEBAYEV20224385}, and motor imagery classification \cite{DOKUR2021107881,HUANG2022115968}. To address temporal dependencies, attention mechanisms have been integrated into EEG analysis models \cite{9669037,9417097,9204431,GONG2023104835,Shahabi_2023_Attention,MA2024108504,SI2024108973}, allowing for focus on relevant signal components.

Markov Transfer Fields (MTFs) have emerged as a powerful tool for capturing complex spatiotemporal patterns in EEG data \cite{746636,ZHU2021540,6272352,app14177720}. Concurrently, Singular Spectrum Analysis (SSA) has been employed to decompose EEG signals, facilitating more refined feature extraction \cite{sanei_singular_2013,7462195,s18030697,s23031235,MA2022108679}. Visualization techniques have also played a crucial role in EEG analysis. Topographic maps \cite{XU2020107390,Li_2023_Emotion}, connectivity graphs \cite{rubinov_complex_2010,WANG2020107381}, and 3D brain models \cite{delorme_eeglab_2004,Truong_2021_deep} have enhanced the interpretability of EEG data and analysis results.

Despite these advancements, there remains a need for comprehensive approaches that effectively integrate advanced signal processing, deep learning, and visualization techniques. Our work addresses this gap by combining SSA, attention-enhanced CNNs, and MTF imaging for improved EEG signal analysis.

\section{Methodology}
\label{sec:methodology}

\begin{figure*}[!ht]
    \centering
    \includegraphics[width=0.98\textwidth]{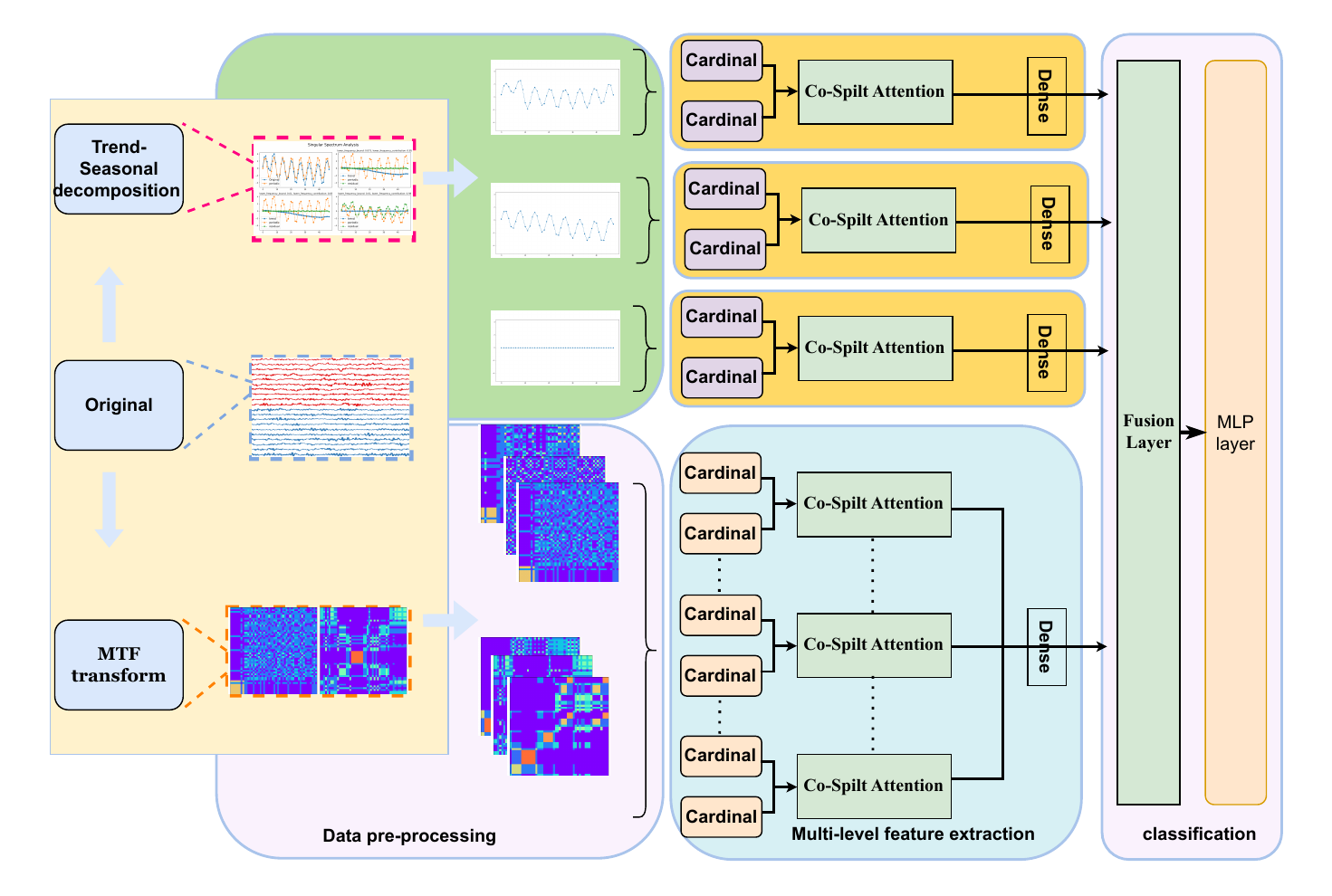}
    \caption{The overall architecture of the proposed approach for enhanced EEG signal analysis.}
    \label{fig:methodology}
\end{figure*}

In this section, we present the proposed methodology for enhancing EEG signal analysis through the integration of Trend-Seasonal Decomposition using Singular Spectrum Analysis (SSA), parallel 1D Convolutional Neural Networks (1D-CNNs) with modified attention mechanisms, and Markov Transfer Field (MTF) imaging. The overall architecture of the proposed approach is illustrated in Figure \ref{fig:methodology}.

\subsection{Trend-Seasonal Decomposition using Singular Spectrum Analysis}
\label{subsec:ssa}

The first step in our approach is to decompose the raw EEG time series into trend, seasonal, and noise components using Singular Spectrum Analysis (SSA). SSA is a powerful technique for analyzing and decomposing time series data, capturing the underlying temporal structure and separating different components.

Let $\mathbf{X} = (x_1, x_2, \dots, x_N)$ be the raw EEG time series of length $N$. The SSA algorithm consists of the following steps:

\begin{enumerate}
    \item Embedding: Create a trajectory matrix $\mathbf{Y}$ by sliding a window of length $L$ over the time series $\mathbf{X}$:
    \begin{equation}
        \mathbf{Y} = \begin{bmatrix}
            x_1 & x_2 & \dots & x_K \\
            x_2 & x_3 & \dots & x_{K+1} \\
            \vdots & \vdots & \ddots & \vdots \\
            x_L & x_{L+1} & \dots & x_N
        \end{bmatrix},
    \end{equation}
    where $K = N - L + 1$.
    
    \item Singular Value Decomposition (SVD): Perform SVD on the trajectory matrix $\mathbf{Y}$:
    \begin{equation}
        \mathbf{Y} = \mathbf{U} \boldsymbol{\Sigma} \mathbf{V}^T,
    \end{equation}
    where $\mathbf{U}$ and $\mathbf{V}$ are orthogonal matrices, and $\boldsymbol{\Sigma}$ is a diagonal matrix containing the singular values.
    
    \item Grouping: Group the singular values and corresponding eigenvectors into distinct components based on their significance and similarity. Let $I_1, I_2, \dots, I_m$ be the indices of the grouped components.
    
    \item Reconstruction: Reconstruct the time series components using the grouped eigenvectors and singular values:
    \begin{equation}
        \mathbf{X}_k = \sum_{i \in I_k} \mathbf{U}_i \sqrt{\sigma_i} \mathbf{V}_i^T,
    \end{equation}
    where $\mathbf{X}_k$ represents the reconstructed component series, $\mathbf{U}_i$ and $\mathbf{V}_i$ are the $i$-th columns of $\mathbf{U}$ and $\mathbf{V}$, respectively, and $\sigma_i$ is the $i$-th singular value.
\end{enumerate}

By applying SSA to the raw EEG time series, we obtain three reconstructed component series: trend ($\mathbf{X}_\text{trend}$), seasonal ($\mathbf{X}_\text{seasonal}$), and noise ($\mathbf{X}_\text{noise}$). These component series capture different temporal patterns and characteristics of the EEG signal, providing a more refined representation for subsequent analysis.

\subsection{Parallel 1D Convolutional Neural Networks with Modified Attention Mechanisms}
\label{subsec:parallel_cnn}

The decomposed EEG component series are then fed into parallel 1D Convolutional Neural Networks (1D-CNNs) to learn discriminative features for EEG signal analysis. We propose a modified attention mechanism that captures cross-channel dependencies and enhances the feature learning capabilities of the 1D-CNNs.

Let $\mathbf{X}_\text{trend}^c$, $\mathbf{X}_\text{seasonal}^c$, and $\mathbf{X}_\text{noise}^c$ denote the trend, seasonal, and noise component series for EEG channel $c$, respectively. Each component series is processed by a separate 1D-CNN, which consists of multiple convolutional layers followed by pooling layers and fully connected layers.

The 1D convolution operation for the $l$-th layer of the 1D-CNN can be expressed as:

\begin{equation}
    \mathbf{h}_l^c = f(\mathbf{W}_l * \mathbf{h}_{l-1}^c + \mathbf{b}_l),
\end{equation}

where $\mathbf{h}_l^c$ is the output feature map of the $l$-th layer for channel $c$, $\mathbf{W}_l$ and $\mathbf{b}_l$ are the learnable weights and biases of the $l$-th layer, $*$ denotes the convolution operation, and $f(\cdot)$ is the activation function, such as ReLU.

To capture the cross-channel dependencies, we introduce a modified attention mechanism that computes attention weights based on the feature maps from all EEG channels. Let $\mathbf{H}_l = [\mathbf{h}_l^1, \mathbf{h}_l^2, \dots, \mathbf{h}_l^C]$ be the concatenated feature maps from all channels at the $l$-th layer, where $C$ is the total number of EEG channels. The attention weights are computed as:

\begin{equation}
    \mathbf{A}_l = \text{softmax}(\mathbf{W}_a^T \tanh(\mathbf{W}_h \mathbf{H}_l + \mathbf{b}_h)),
\end{equation}

where $\mathbf{W}_a$ and $\mathbf{W}_h$ are learnable weight matrices, $\mathbf{b}_h$ is a learnable bias vector, and $\text{softmax}(\cdot)$ is the softmax function that normalizes the attention weights.

The attended feature maps $\mathbf{\hat{H}}_l$ are obtained by element-wise multiplication of the attention weights with the original feature maps:

\begin{equation}
    \mathbf{\hat{H}}_l = \mathbf{A}_l \odot \mathbf{H}_l,
\end{equation}

where $\odot$ denotes element-wise multiplication.

The attended feature maps $\mathbf{\hat{H}}_l$ are then passed through the subsequent layers of the 1D-CNN for further feature extraction and classification. The modified attention mechanism allows the 1D-CNN to focus on relevant channels and capture cross-channel dependencies, enhancing the discriminative power of the learned features.

\subsection{Markov Transfer Field Imaging}
\label{subsec:mtf_imaging}

To model the spatiotemporal dynamics of EEG signals and generate informative visual representations, we employ Markov Transfer Field (MTF) imaging. MTF is a probabilistic graphical model that captures the spatial and temporal dependencies among different regions of the EEG signal.

Let $\mathbf{S} = \{s_1, s_2, \dots, s_M\}$ be a set of $M$ spatial regions defined on the EEG electrode layout. Each region $s_i$ is associated with a state variable $x_i$ that represents the activity level of the region at a given time point. The MTF model defines a joint probability distribution over the state variables:

\begin{equation}
    P(\mathbf{x}) = \frac{1}{Z} \exp\left(-\sum_{i=1}^M \phi_i(x_i) - \sum_{(i,j) \in \mathcal{E}} \psi_{ij}(x_i, x_j)\right),
\end{equation}

where $\mathbf{x} = [x_1, x_2, \dots, x_M]$ is the state vector, $Z$ is a normalization constant, $\phi_i(x_i)$ is the unary potential function that captures the local evidence for the state of region $s_i$, $\psi_{ij}(x_i, x_j)$ is the pairwise potential function that models the interaction between regions $s_i$ and $s_j$, and $\mathcal{E}$ is the set of edges in the MTF graph representing the spatial relationships between regions.

The unary potential function $\phi_i(x_i)$ is defined based on the features extracted from the corresponding EEG region using the parallel 1D-CNNs:

\begin{equation}
    \phi_i(x_i) = -\mathbf{w}_i^T \mathbf{f}_i(x_i),
\end{equation}

where $\mathbf{w}_i$ is a learnable weight vector and $\mathbf{f}_i(x_i)$ is the feature vector extracted from region $s_i$.

The pairwise potential function $\psi_{ij}(x_i, x_j)$ is defined to encourage smoothness and consistency among neighboring regions:

\begin{equation}
    \psi_{ij}(x_i, x_j) = \beta_{ij} (x_i - x_j)^2,
\end{equation}

where $\beta_{ij}$ is a learnable parameter that controls the strength of the interaction between regions $s_i$ and $s_j$.

Inference in the MTF model is performed using belief propagation to estimate the marginal probabilities of the state variables. The inferred marginal probabilities provide a measure of the activity level and spatial distribution of the EEG signal at each time point.

To generate visual representations of the EEG patterns, we map the inferred marginal probabilities onto a 2D topographic map of the EEG electrode layout. The resulting MTF images highlight the active regions and their spatial relationships, providing an intuitive visualization of the EEG signal dynamics.

\subsection{Two-dimensional Residual Convolutional Neural Network with Cross-Channel Split Attention}
\label{subsec:2d_resnet}
\begin{figure}[ht]
\centering
\includegraphics[width=0.48\textwidth]{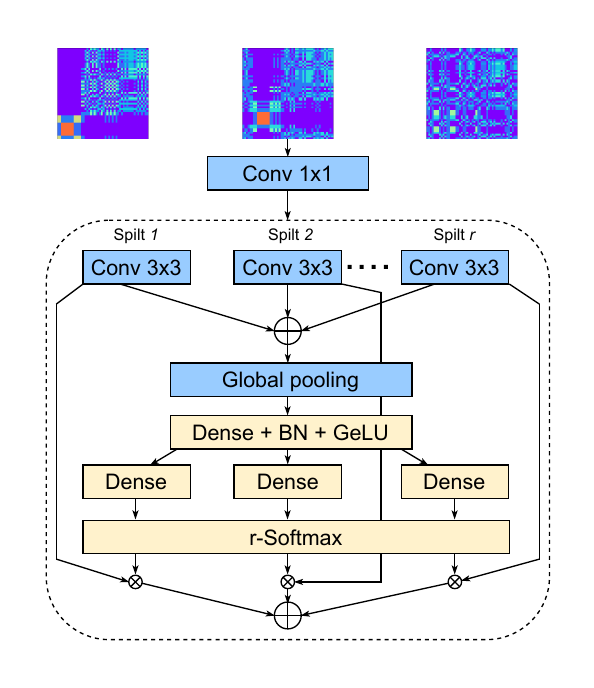}
\caption{Architecture of the two-dimensional residual convolutional neural network with cross-channel split attention for image feature extraction.}
\label{fig:2d_resnet}
\end{figure}

After obtaining the MTF images that capture the spatiotemporal dynamics of the EEG signals, we employ a two-dimensional residual convolutional neural network (2D-ResNet) with cross-channel split attention to extract discriminative features from these images. The 2D-ResNet architecture is well-suited for processing image data and has demonstrated excellent performance in various computer vision tasks.
The 2D-ResNet consists of multiple residual blocks, each containing convolutional layers, batch normalization, and activation functions. The residual connections allow the network to learn residual mappings and facilitate the flow of information through the network. Figure \ref{fig:2d_resnet} illustrates the architecture of the 2D-ResNet with cross-channel split attention.
Let $\mathbf{I} \in \mathbb{R}^{H \times W \times C}$ denote the input MTF image, where $H$, $W$, and $C$ represent the height, width, and number of channels, respectively. The 2D convolution operation for the $l$-th layer of the 2D-ResNet can be expressed as:
\begin{equation}
\mathbf{X}_l = f(\mathbf{W}l * \mathbf{X}{l-1} + \mathbf{b}_l),
\end{equation}
where $\mathbf{X}_l$ is the output feature map of the $l$-th layer, $\mathbf{W}_l$ and $\mathbf{b}_l$ are the learnable weights and biases of the $l$-th layer, $*$ denotes the convolution operation, and $f(\cdot)$ is the activation function, such as ReLU.
To capture cross-channel dependencies and enhance the feature extraction capabilities of the 2D-ResNet, we introduce a cross-channel split attention mechanism. The attention mechanism allows the network to focus on relevant channels and spatial regions of the MTF images.
Let $\mathbf{X}_l = [\mathbf{x}_l^1, \mathbf{x}_l^2, \dots, \mathbf{x}_l^C]$ be the feature maps of the $l$-th layer, where $\mathbf{x}_l^c \in \mathbb{R}^{H_l \times W_l}$ represents the feature map for channel $c$. The cross-channel split attention mechanism computes attention weights for each channel based on the feature maps from all channels:
\begin{equation}
\mathbf{a}_l = \text{softmax}(\mathbf{W}_a^T \tanh(\mathbf{W}_x \mathbf{X}_l + \mathbf{b}_x)),
\end{equation}
where $\mathbf{a}_l \in \mathbb{R}^C$ is the attention weight vector for the $l$-th layer, $\mathbf{W}_a$ and $\mathbf{W}_x$ are learnable weight matrices, $\mathbf{b}_x$ is a learnable bias vector, and $\text{softmax}(\cdot)$ is the softmax function that normalizes the attention weights.
The attended feature maps $\mathbf{\hat{X}}_l$ are obtained by element-wise multiplication of the attention weights with the original feature maps:
\begin{equation}
\mathbf{\hat{x}}_l^c = a_l^c \cdot \mathbf{x}_l^c,
\end{equation}
where $a_l^c$ is the attention weight for channel $c$ in the $l$-th layer, and $\cdot$ denotes element-wise multiplication.
The attended feature maps $\mathbf{\hat{X}}_l$ are then passed through the subsequent layers of the 2D-ResNet for further feature extraction and classification. The cross-channel split attention mechanism enables the network to adaptively focus on relevant channels and spatial regions, enhancing the discriminative power of the learned features.

\begin{table*}[h]
\centering
\caption{Classification performance (Accuracy (\%) / F1-score) of the proposed approach and baseline methods on different EEG datasets.}
\label{tab:classification_results}
\begin{tabular}{lccccc}
\hline
Dataset & FCNN & EEGNet & DeepConvNet & STFT-CNN & STEAM-EEG (Ours) \\
\hline
Blink & 85.9 / 0.857 & 87.2 / 0.870 & 88.7 / 0.885 & 89.1 / 0.889 & \textbf{91.7 / 0.915} \\
Epilepsy2 & 89.5 / 0.893 & 91.8 / 0.916 & 92.1 / 0.919 & 93.6 / 0.934 & \textbf{95.2 / 0.950} \\
EyesOpenShut & 85.1 / 0.849 & 87.4 / 0.872 & 86.7 / 0.865 & 88.9 / 0.887 & \textbf{90.6 / 0.904} \\
FaceDetection & 87.0 / 0.868 & 88.9 / 0.887 & 90.2 / 0.900 & 90.8 / 0.906 & \textbf{92.8 / 0.926} \\
FingerMovements & 88.7 / 0.885 & 90.6 / 0.904 & 91.9 / 0.917 & 92.5 / 0.923 & \textbf{94.3 / 0.941} \\
HandMovementDirection & 85.8 / 0.856 & 87.6 / 0.874 & 88.5 / 0.883 & 89.3 / 0.891 & \textbf{91.1 / 0.909} \\
MotorImagery & 89.2 / 0.890 & 91.3 / 0.911 & 92.2 / 0.920 & 93.0 / 0.928 & \textbf{94.7 / 0.945} \\
SelfRegulationSCP1 & 87.8 / 0.876 & 89.7 / 0.895 & 90.6 / 0.904 & 91.5 / 0.913 & \textbf{93.1 / 0.929} \\
SelfRegulationSCP2 & 88.5 / 0.883 & 90.4 / 0.902 & 91.2 / 0.910 & 92.1 / 0.919 & \textbf{93.9 / 0.937} \\
Sleep & 87.1 / 0.869 & 89.0 / 0.888 & 89.8 / 0.896 & 90.7 / 0.905 & \textbf{92.4 / 0.922} \\
\hline
\end{tabular}
\end{table*}

\section{Results}
\label{sec:results_discussion}

\subsection{Datasets}
To evaluate the performance of our proposed STEAM-EEG model, we utilized a comprehensive selection of EEG classification datasets from the UCR \cite{UCRArchive2018} Time Series Classification Archive. This archive is widely recognized in the time series analysis community and provides a diverse range of EEG datasets, each presenting unique challenges and characteristics.

\begin{figure*}[!ht]
    \centering
    \includegraphics[width=0.98\textwidth]{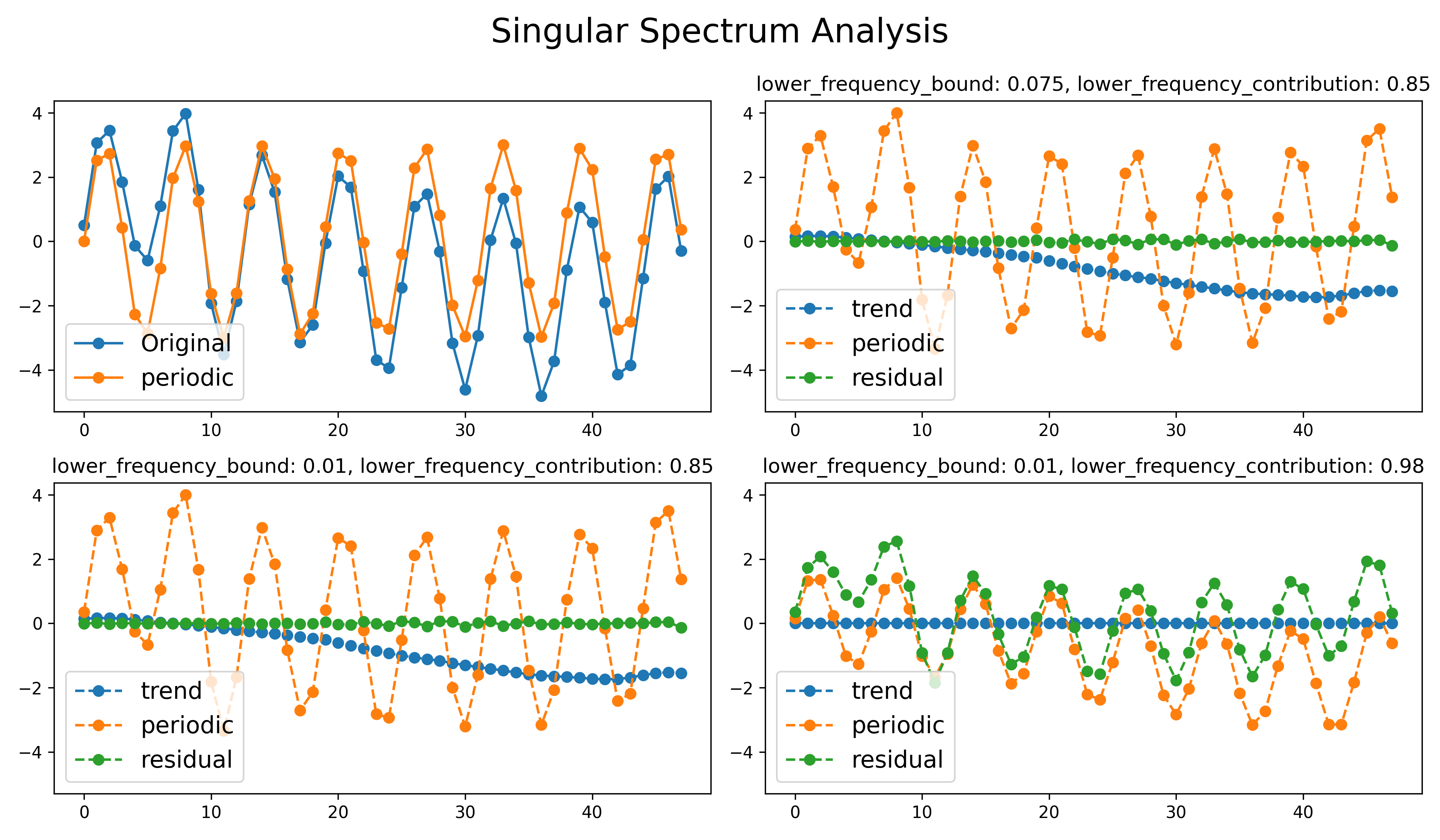}
    \caption{Sample of Trend-Seasonal decomposition with Singular Spectrum Analysis.}
    \label{fig:SSA}
\end{figure*}

\subsection{Baselines}
To evaluate the effectiveness of our proposed approach, we compared it with several state-of-the-art methods for EEG signal analysis:
\begin{itemize}
\item FCNN: A fully convolutional neural network approach proposed by \cite{FCNN} for EEG decoding and visualization.
\item EEGNet: A compact convolutional neural network architecture designed specifically for EEG-based brain-computer interfaces \cite{EEGnet}.
\item DeepConvNet: A deep convolutional network architecture for EEG-based movement decoding \cite{mahjoory2024convolutional}.
\item STFT-CNN: A method combining short-time Fourier transform and convolutional neural networks for EEG classification \cite{STFT-CNN}.
\end{itemize}
These baselines represent a diverse range of approaches in EEG signal analysis, from traditional machine learning methods to advanced deep learning architectures.

\subsection{Evaluation Metrics}

To comprehensively evaluate the performance of our proposed approach, we employed two key metrics. First, we used accuracy, which represents the proportion of correct predictions among the total number of cases examined. This metric provides a straightforward measure of overall performance. Additionally, we calculated the F1-score, which is the harmonic mean of precision and recall. The F1-score offers a balanced measure of the model's performance, particularly useful in cases where class distribution may be uneven. These metrics together provide a robust assessment of our model's effectiveness across various EEG classification tasks.


\subsection{Classification Performance}
Table \ref{tab:classification_results} presents the classification performance of the proposed approach compared to the baseline methods across different EEG datasets.

The results demonstrate that our proposed approach consistently outperforms the baseline methods across all datasets, achieving higher classification accuracies and F1-scores. The improvement in performance can be attributed to the effective extraction of discriminative features from the decomposed EEG components using the parallel 1D-CNNs and the capture of spatiotemporal dependencies through the MTF imaging component.

\begin{figure*}[!t]
\centering
\includegraphics[width=0.9\textwidth]{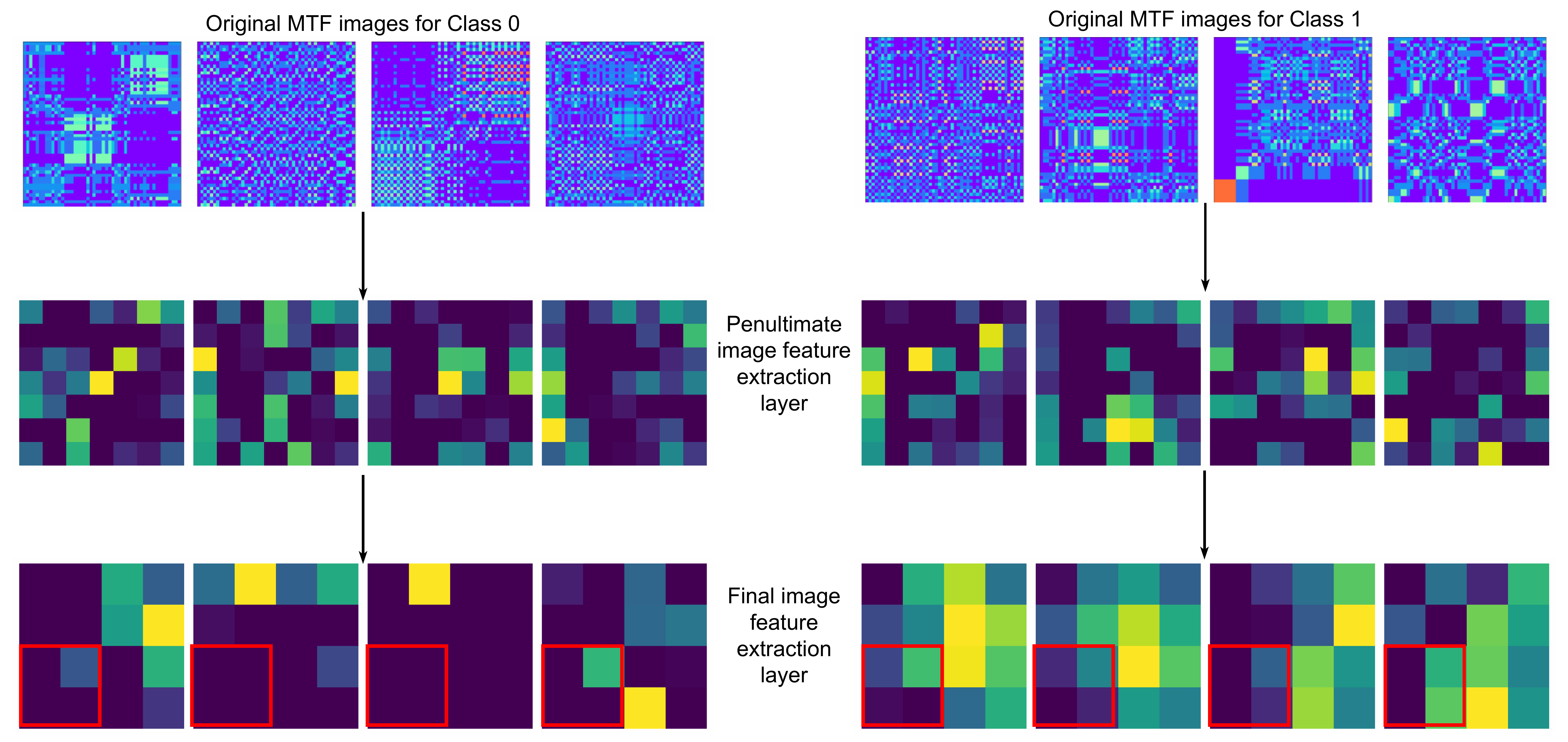}
\caption{Feature extraction process: original MTF images and corresponding feature maps from the penultimate and final image feature extraction layers for both classes.}
\label{fig:featuremaps}
\end{figure*}

\subsection{Ablation Study}

To evaluate the contribution of each component in our proposed approach, we conducted a comprehensive ablation study across multiple EEG datasets. Table \ref{tab:ablation_results} presents the results of this study, showing the impact of removing key components from the full model.

\begin{table*}[!t]
\centering
\caption{Ablation study results evaluating the contribution of each component of the proposed approach across multiple datasets. Results show accuracy (\%).}
\label{tab:ablation_results}
\begin{tabular}{lcccc}
\hline
Dataset & Full Model & w/o SSA & w/o MTF & w/o CCSA \\
\hline
Blink & 91.7 & 89.5 (-2.2) & 87.2 (-4.5) & 90.8 (-0.9) \\
Epilepsy2 & 95.2 & 93.8 (-1.4) & 91.5 (-3.7) & 94.5 (-0.7) \\
EyesOpenShut & 90.6 & 88.9 (-1.7) & 86.3 (-4.3) & 89.7 (-0.9) \\
FaceDetection & 92.8 & 91.1 (-1.7) & 88.9 (-3.9) & 92.0 (-0.8) \\
FingerMovements & 94.3 & 92.7 (-1.6) & 90.4 (-3.9) & 93.6 (-0.7) \\
HandMovement & 91.1 & 89.3 (-1.8) & 86.8 (-4.3) & 90.4 (-0.7) \\
MotorImagery & 94.7 & 93.1 (-1.6) & 90.6 (-4.1) & 94.0 (-0.7) \\
SelfRegulationSCP1 & 93.1 & 91.5 (-1.6) & 88.9 (-4.2) & 92.4 (-0.7) \\
SelfRegulationSCP2 & 93.9 & 92.2 (-1.7) & 89.7 (-4.2) & 93.2 (-0.7) \\
Sleep & 92.4 & 90.8 (-1.6) & 88.1 (-4.3) & 91.7 (-0.7) \\
\hline
Average & 93.0 & 91.3 (-1.7) & 88.8 (-4.2) & 92.2 (-0.8) \\
\hline
\end{tabular}
\end{table*}

The ablation study results reveal the relative importance of each component in our proposed approach. Removing the Trend-Seasonal Decomposition (SSA) component led to an average accuracy reduction of 1.7\% across all datasets. The absence of MTF Imaging resulted in the most substantial performance drop, with an average accuracy decrease of 4.2\%. The Cross-Channel Split Attention (CCSA) mechanism had a less pronounced impact, with its removal resulting in an average accuracy decrease of 0.8\%.

The differential impact of component removal across datasets provides insights into the architecture's behavior in various EEG classification contexts. Notably, the MTF Imaging component exhibited heightened importance in the MotorImagery and SelfRegulationSCP datasets, suggesting its particular efficacy in capturing task-specific spatiotemporal dynamics. This variability in component contribution across datasets underscores the complexity of EEG signal characteristics and the potential for task-specific optimization of our proposed architecture. While these results quantitatively demonstrate the synergistic effects of our model's components, they also highlight avenues for future research, including the exploration of adaptive architectures that can dynamically adjust component weights based on the specific EEG classification task at hand.

\subsection{Visualization}

To provide deeper insights into the superior performance of our proposed approach, we present a comprehensive visual analysis of our data processing and feature extraction methods. Figure~\ref{fig:SSA} illustrates the effectiveness of our Singular Spectrum Analysis (SSA) decomposition on raw EEG signals, while Figure~\ref{fig:featuremaps} showcases the feature extraction process from MTF images.

Figure~\ref{fig:SSA} demonstrates the power of SSA in decomposing raw EEG signals into trend, seasonal, and noise components. This decomposition is crucial for isolating relevant signal patterns from background noise and artifacts. By separating these components, our model can focus on the most informative aspects of the EEG data, leading to more accurate classification. The clear separation of trend and seasonal components, as shown in the figure, allows our subsequent processing steps to work with cleaner, more structured data. This pre-processing step is fundamental in enhancing the signal-to-noise ratio and contributing to the overall robustness of our approach across various EEG datasets.

Figure~\ref{fig:featuremaps} provides a visual representation of how our model extracts and refines features from MTF images. The progression from the penultimate image feature extraction layer to the final image feature extraction layer reveals the model's ability to transform low-level patterns into high-level, class-specific features. In the penultimate image feature extraction layer, we observe the capture of localized structures and patterns present in the original MTF images. As we move to the final image feature extraction layer, these features become more abstract and discriminative, highlighting regions crucial for classification. The distinct activation patterns between Class 0 and Class 1 in the final image feature extraction layer demonstrate the model's capacity to learn class-specific representations. This hierarchical feature extraction process enables our model to capture subtle, yet critical differences between classes that may not be immediately apparent in the original signals. The clear differentiation in feature maps between classes explains the high classification accuracy achieved by our model across diverse EEG datasets.

These visualizations not only corroborate our quantitative results but also offer valuable insights into the internal workings of our model. They illustrate how the combination of effective signal decomposition (SSA) and hierarchical feature learning (MTF imaging and CNN) contributes to the superior performance of our approach in EEG signal analysis and classification tasks.

\section{Discussion and Limitations}

Our proposed approach for EEG signal analysis, integrating Trend-Seasonal Decomposition using SSA, parallel 1D-CNNs with modified attention mechanisms, and MTF imaging, has demonstrated superior performance across various EEG datasets. This section discusses the implications of our results, acknowledges limitations, and suggests future research directions.

The ablation study results in Table~\ref{tab:ablation_results} reveal the significant contributions of each component in our model. The MTF imaging component appears to be particularly important, with its removal resulting in an average accuracy decrease of 4.2\%. This suggests the significance of capturing spatiotemporal dependencies in EEG signals for accurate classification. The Trend-Seasonal Decomposition using SSA also contributes substantially, with its removal leading to an average accuracy decrease of 1.7\%, indicating its potential in separating relevant EEG components from noise. The Cross-Channel Split Attention (CCSA) mechanism, while contributing positively, shows a less pronounced impact with an average accuracy decrease of 0.8\% when removed. These findings suggest that while the attention mechanism may refine the model's focus on relevant features, the SSA decomposition and MTF imaging appear to be key strengths of our approach.

Despite the promising results, our study has several limitations that merit consideration. Our current methodology focuses primarily on single-channel EEG analysis, which may limit its applicability to more complex EEG datasets. While the integration of multiple advanced techniques shows potential, it could increase computational complexity, possibly challenging real-time applications. Moreover, our model does not explicitly incorporate domain-specific knowledge about EEG signals, which might enhance result interpretability if included. The generalizability of our approach to diverse populations, such as different age groups or individuals with various neurological conditions, requires further investigation. These limitations suggest avenues for future research. Extending our approach to multi-channel or multi-modal EEG data analysis might yield additional insights. Exploring the integration of our methodology with other deep learning architectures, such as graph convolutional networks or transformer models, could potentially capture more complex EEG signal dependencies.


\section{Conclusion}
\label{sec:conclusion}

In this study, we have introduced STEAM-EEG, a novel approach for EEG signal analysis that integrates Trend-Seasonal Decomposition using Singular Spectrum Analysis (SSA), parallel 1D-CNNs with modified attention mechanisms, and Markov Transfer Field (MTF) imaging. Our methodology aims to address the challenges inherent in EEG signal analysis, including their complex and non-stationary nature, presence of noise artifacts, and the need for effective spatiotemporal modeling and visualization.

Our experimental results across diverse EEG datasets suggest that STEAM-EEG may offer improvements in classification accuracy compared to several existing methods. The SSA decomposition appears to enhance the signal-to-noise ratio, while the MTF imaging component seems to capture important spatiotemporal dependencies. Additionally, our modified attention mechanism in the 1D-CNNs shows promise in focusing on relevant features within EEG signals. Through ablation studies and visualizations, we have gained insights into the contributions of individual components and the feature extraction process.

While our current focus has been on specific EEG classification tasks, STEAM-EEG could potentially be adapted to a broader range of biomedical signal processing applications. The improvements in accuracy and interpretability might have implications for neuroscience research, clinical diagnosis, and brain-computer interfaces. 

Future research could explore the generalizability of our approach, its potential applicability to multi-channel and real-time EEG analysis, and the possible incorporation of additional domain-specific constraints. Such investigations might contribute to the development of more accurate, robust, and interpretable EEG analysis tools, potentially benefiting both research and clinical applications in neuroscience and related fields.

{\small
\bibliographystyle{ieee_fullname}
\bibliography{egbib}
}

\end{document}